\documentclass[letterpaper, 10 pt, conference]{ieeeconf}  
\usepackage{graphicx}
\usepackage{booktabs, caption, makecell}

\usepackage{threeparttable}
\usepackage{dblfloatfix}
\usepackage{here}

\IEEEoverridecommandlockouts                              

\title{\LARGE \bf
Empirical Investigation of Factors that Influence\\ Human Presence and Agency in Telepresence Robot
}
\author{Nungduk Yun$^{1}$ and Seiji Yamada$^{2}$
	\thanks{$^{1}$Nungduk Yun is with The Graduate University for Advance Studies,
		SOKENDAI,
		Tokyo, Japan
		{\tt\small ndyun@nii.ac.jp}}%
	\thanks{$^{2}$ Seiji Yamada with National Institute of Informatics, NII, Tokyo, Japan,
		{\tt\small seiji@nii.ac.jp}}%
}

\begin{document}

\maketitle
\thispagestyle{empty}
\pagestyle{empty}

\begin{abstract}
Nowadays, a community starts to find the need for human presence in an alternative way, there has been tremendous research and development in advancing telepresence robots. People tend to feel closer and more comfortable with telepresence robots as many senses a human presence in robots. In general, many people feel the sense of agency from the face of a robot, but some telepresence robots without arm and body motions tend to give a sense of human presence. It is important to identify and configure how the telepresence robots affect a sense of presence and agency to people by including human face and slight face and arm motions. Therefore, we carried out extensive research via web-based experiment to determine the prototype that can result in soothing human interaction with the robot. The experiments featured videos of a telepresence robot ($\textbf{n}$ = 128, $\textbf{2}\times\textbf{2}$ between-participant study robot face factor: video-conference, robot-like face; arm motion factor: moving vs. static) to investigate the factors significantly affecting human presence and agency with the robot. We used two telepresence robots: an affordable robot platform and a modified version for human interaction enhancements. The findings suggest that participants feel agency that is closer to human-likeness when the robot's face was replaced with a human's face and without a motion. The robot's motion invokes a feeling of human presence whether the face is human or robot-like.
\end{abstract}
\section{INTRODUCTION}
There is a wide variety of means of digital communication. For example, the telephone, e-mail, SNS, and video conferencing are representative and the most common. In recent years, communication via telepresence robots has been attracting attention. Telepresence robots called ``mobile robotic presence systems,'' that is physical robotic platforms with a video-conferencing system mounted on a robotic mobile platform \cite{Kristoffersson2013}.
Since remote operators can control this physical embodiment which is telepresence robot in a remote location, these systems chance to increase for video-conference and audio-communication \cite{Kristoffersson2013}\cite{Rae2013} and imbue communicators with a strong sense of presence \cite{Rae2014}. Other researchers have mentioned that telepresence robots provide a new communication platform in various situations. For example, business, education, and Medical related \cite{Lee2011}\cite{Tsui2011}\cite{Neustaedter2016}\cite{Fels2001}\cite{movement17}. 
Furthermore, recently, these robots have become a new platform for people with disabilities to use so that they can join in society and perform labor \cite{Takeuchi2020}. 

In the field of telepresence robots, it is said that the visibility of a face is a factor that makes people feel the presence of a person \cite{Rae2014}\cite{movement17}.  
Previous research on movement with telepresence robots has shown that robots that can express themselves socially through movement are more immersive and desirable than those that do not move \cite{Adalgeirsson2010}. In some cases, a robot that does not have a human face can make people feel the presence of a person through its movements \cite{Orylab}. For telepresence robots, the effect the face has depends on whether the face is human-like or robot-like, and people may feel that one type gives a greater sense of agency or presence. Also, which makes people feel more agency or presence in a robot, the case with or without arm motion which is moving or static? 

In this paper, we conducted a web-based experiment featuring video of a telepresence robot ($n$ = 128, $2\times2$ between-participant study, robot face factor: video-conference, robot-like face; arm motion: moving, static) to investigate which factors significantly affect human presence and agency.
\section{related work}
\subsection{Robots' Facial Cues in HRI}

One of the most important factors in human-robot interaction (HRI) is the ability to show facial cues. In literature, McGinn conducted various studies on service robots by changing their heads and facial cues to determine the effect on social interface between human and robots. To facilitate better communication between human and robot, it is common for robots to possess head-like features that is capable of providing social feedback. Although it is not a human’s face and expression, research has shown that relation between human and the robot can be enhanced when the robot is equipped with human-like “robotic” face that can express and show motion like humans \cite{McGinn2020a}. In addition, interaction with robot gets even better when the robot exhibits social behaviors with anthropomorphic characteristics \cite{McGinn2020} \cite{McGinn2020a}. 

Effect of human’s facial cue on robots were discussed in Mollahosseini et al.   \cite{Mollahosseini2018}. The study indicated that eye gaze and certain facial expressions from the actual human can further improve the relation between human and robots by physically displaying human’s face on the 2D screen using telecommunication either telepresence or a virtual agents \cite{Mollahosseini2018}. For example, Beam is one of the known telepresence robotic system that utilizes this method by replacing robot’s head with the LED screen to display human’s face via video-conference. Several research and surveys indicated that people can really feel the presence of human in the robot \cite{Neustaedter2016}\cite{Rae2014}. 
\subsection{Non-verbal Cues in HRI}

Previous research indicated that, non-verbal cues are becoming important factors in HRI technology as it plays a significant role in human-human interaction. Telepresence robot that has an ability to show social expression by the motion movement can make user to feel engagement and likeable \cite{Adalgeirsson2010}. OriHime is one of the telepresence and avatar robot with human-like behavior that was designed by Ory Laboratory Inc. The purpose of this technology is to aid people to engage in life-like social behavior over distance, facilitate human-to-human social interaction, and serve as an avatar to the user \cite{Vikkels2020}. Although OriHime does not have an expressive facial cue, but its limb movement and the voice from the user relaxes the social-relation between human and the robot and still feel the presence of human in them \cite{Orylab}. 
In addition, having social expression embodiment like Orhime-D can join society and social events for disabled people and fulfill their mentality \cite{Takeuchi2020}. Spatial configuration and body orientation of a telepresence robot affected people enable to arrange themselves, robot tend to copy human-like action and theyt detected by the surrounded motion \cite{Kuzuoka2010}. This greatly increases interaction quality between human and robots.

In the research using teleoperated robot with the Wizard of OZ method, participants was not affected enjoyment by the knowledge of whether the robot was being controlled by a program or a human \cite{Yamaoka2007}. Yamada et al. proposed motion-based ASE(Artificial Subtle Expressions) in which a robot slowly hesitates by turning to a human before giving advice with low confidence.  As they conclude, the long or short-wait expressions might be applicable to expressing a robot's confidence, fast or slow-motion as a motion-ASE is more suitable for such expressions \cite{Yamada2013}. The other research suggest that synchronized on-screen and in-space gestures significantly improved viewers(participants)’ interpretation of the action compared to on-screen or in-space gestures alone and addition of proxy motion also improved measures of perceived collaboration \cite{Sirkin2012}. Furthermore, in-space gestures positively influenced perceptions of both local and remote participants \cite{Sirkin2012}. 


\section{Platform}
We used a humanoid robot, Rapiro \cite{rapiro}, which is widely used for different applications, such as for education and hobbies\cite{Abiri2015}\cite{DAuria2018}. The Arduino and Raspberry Pi boards in the robot enable users (developers) to communicate with the robot by only sending command signals from a PC, and they also allow for the system to be extended easily. Therefore, we used this robot as a telepresence robot for our experiment. 

For the experiment, we fixed Rapiro's eye color to blue due to color bias. We modified Rapiro's head as a prototype for experimental study and to show the remote user's face. 
\subsection{Hardware Spec}
Rapiro \footnote[1]{http://www.rapiro.com/} has 12 degrees of freedom (DoF), a USB camera, a microphone in its forehead, and a speaker inside of its head. Fig. \ref{fig:rapiro3} shows an overview of Rapiro.
	
Modifications to Rapiro, we modified the head of another Rapiro to show the face of a remote user. We used a 5-inch portable monitor, and the head was made of PLA using a 3D printer. This Rapiro also had 12 DoF, a USB camera, microphone, and speaker inside of its head. Fig. \ref{fig:rapiro1} shows an overview of the modified Rapiro.
\subsection{User Interface and Robot Motion}
To control Rapiro, we made a keyboard input interface. When the operator (from a geographically separate location) presses the number ``2" on the keyboard, a local PC receives the signal from the operator's location via Wi-Fi, and the robot makes that specific motion. 

We generated robot motions in accordance with the following principle. For both of the robots, we used preset motions and the original motion which we developed for the experiments, such as ``hands up" and ``wave the both hands"and etc. In total, we used six motions in the video task and motions list showed in Table \ref{fig:motionlist} . The preset motions included motion like going forward or backward, but we did not use these at this time.
\begin{table}[ht]
	\centering
	\caption {List of Motions}
	\label{fig:motionlist}
	\begin{tabular}{ll}
		\hline
		& \textbf{Motion}        \\ \hline
		\multicolumn{1}{l|}{1} & Both hands up          \\
		\multicolumn{1}{l|}{2} & Wave the both hands    \\
		\multicolumn{1}{l|}{3} & Stretch our right hand \\
		\multicolumn{1}{l|}{4} & Grip both hands        \\
		\multicolumn{1}{l|}{5} & Left hand up           \\
		\multicolumn{1}{l|}{6} & Fluttering both hands  \\ \hline
	\end{tabular}
\end{table}

\section{Experiment}
\subsection{Experimental Design}
\begin{figure}[ht]
	\begin{tabular}{cc}
		\begin{minipage}[t]{0.5\hsize}
			\includegraphics[width=4cm]{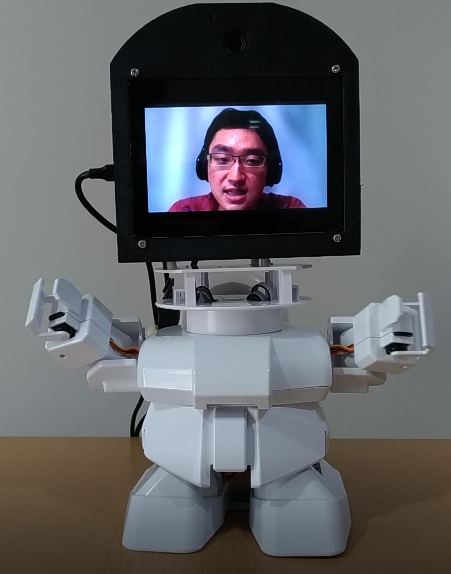}
			\caption{Human face and Moving.}
			\label{fig:rapiro1}
		\end{minipage}
		\begin{minipage}[t]{0.5\hsize}
			\includegraphics[width=3.75cm]{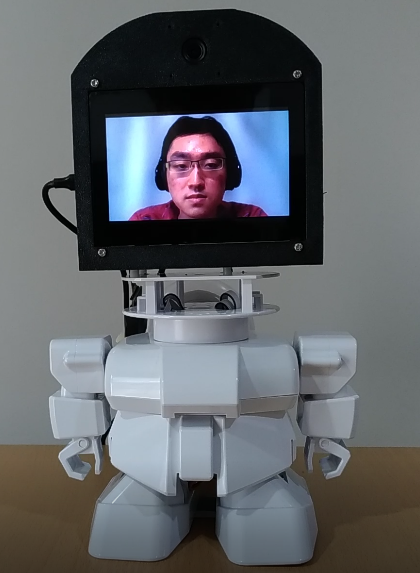}
			\caption{Human face and Static.}
			\label{fig:rapiro2}
		\end{minipage}
		\\
		\begin{minipage}[t]{0.5\hsize}
			\includegraphics[width=3.9cm]{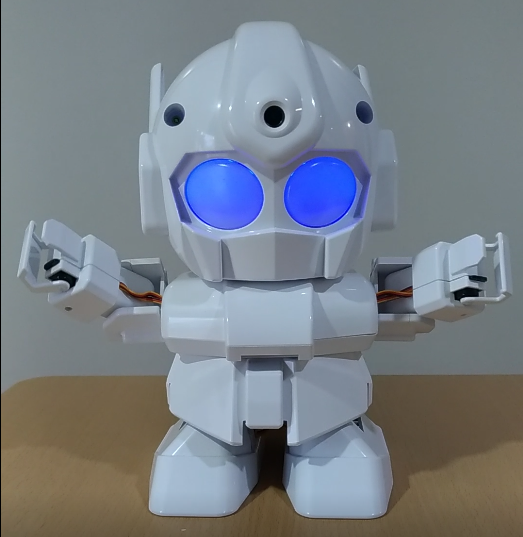}
			\caption{Robot face and Moving.}
			\label{fig:rapiro3}
		\end{minipage}
		\begin{minipage}[t]{0.5\hsize}
			\includegraphics[width=4cm]{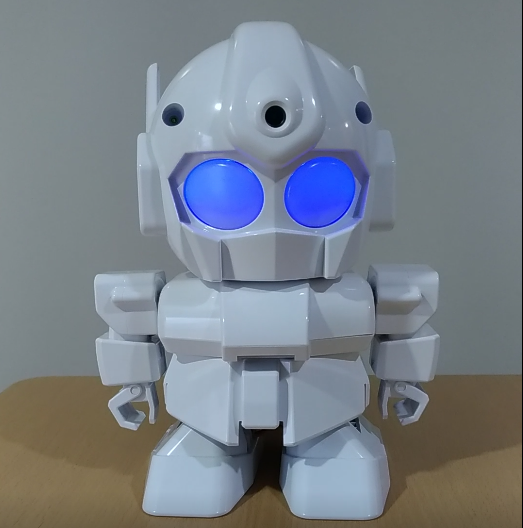}
			\caption{Robot face and Static.}
			\label{fig:rapiro4}
		\end{minipage}
	\end{tabular}
\end{figure}

We conducted the experiment using a $2\times2$ (robot face: video-conference vs. robot-like face; arm motion: moving vs. static) between-participant design. To explore how people interact widely with our design, we used G*Power \cite{Erdfelder2009} sample size calculation (n =128), and we ran our experiment using online questionnaire surveys along with showing the video. Participants were recruited from the Yahoo! Crowdsourcing service and after finished experiments we used Gooogle form for survey. For most of the methods for the online experiment, we referred to Sirkin et al. \cite{Sirkin2012}. 
We wanted the remote operator's speech and gestures to have precise timings and the same interaction content. While online responses may differ from in-person experiments, from Powers et al. ~\cite{Powers2007}, it was found that remote robots could be used in experiments and be more sociable and engaging than co-located robots ~\cite{Powers2007}. Furthermore, studies comparing live and video-based HRI trials were both broadly equivalent in most cases \cite{Woods2006}. Therefore, we chose to run the experiment online. 
Our dependent values were presence and agency. We designed the online experiment so that we could compare which condition affect from our dependent values and perceptions of the remote operator who communicated with them via the telepresence robot across four robot conditions: 
\begin{enumerate}
	
	\item \textbf{Human face and moving}: The face of the robot used a video-conference style screen from which the remote operator was shown, and the robot's arm made motions. An overview of this robot is shown in Fig. \ref{fig:rapiro1}.
	
	\item \textbf{Human face and Static}: The face of the robot used a video-conference style screen from which the remote operator was shown, and the robot's arm was static. An overview is shown in Fig. \ref{fig:rapiro2}.
	
	\item \textbf{Robot face and moving}: The face of the robot was robot-like, and the robot's arm made motions. An overview is shown in Fig. \ref{fig:rapiro3}.
	
	\item \textbf{Robot face and Static}: The face of the robot was robot-like, and the robot's arm was static. An overview is shown in Fig. \ref{fig:rapiro4}.
	
\end{enumerate}
\subsection{Hypotheses}
We formulated four hypotheses for our experiment. As mentioned above, we conducted the experiment using a between-participant design (robot face: video-conference vs. robot-like face; arm motion: moving vs. static) to investigate which factors significantly affect presence and agency.
\begin{itemize}
	\item \textbf{H1} Face affects agency.
	\item \textbf{H2} Motion affects agency.
	\item \textbf{H3} Face affects presence.
	\item \textbf{H4} Motion affects presence.

\end{itemize}
\subsection{Participants}
A total of 216 participants took part in the experiment online (male: 147, female: 69). Their ages ranged from 18 to 63 (M = 44.2, SD = 10.6). We recruited the participants from Yahoo! Crowdsourcing, which is a service provided by Yahoo! Japan.
\subsection{Task} 
The participants watched one video from among the four different conditions as shown in Fig.1,2,3,4. We created videos in which a remote operator communicated via a telepresence robot and discussed moon survival and item ranking. We created this Moon Survival scenario from the Desert Survival Problem \cite{lafferty74} and also a NASA exercise \cite{moon}  since the Desert Survival Problem is used by many social scientists and robotics researchers \cite{Adalgeirsson2010}\cite{Biocca2001}\cite{movement17}\cite{Rae2013}\cite{Yonezu2017}. The video was about an astronaut who had crash-landed on the moon and was discussing how to select 5 items that he needed from the 15 items left to return to his distant home planet. Due to the video length, we only discussed ranking up to five items because if we had gone up to 15 items, the video length would have been too long. The ranking of the five items is shown in the Table \ref{fig:rank of items}. 
\begin{table}[ht]
	\centering
	\caption {Rank of Items}
	\label{fig:rank of items}
	\begin{tabular}{|l|l|l|}
		\hline
		& \textbf{experimenter} & \textbf{Remote operator} \\ \hline
		1 & Tanks of oxygen                & Tanks of oxygen                     \\ \hline
		2 & Map of the Moon's surface                  & Water                        \\ \hline
		3 & Nylon rope               & Food concentrate                    \\ \hline
		4 & Parachute                & First aid kit            \\ \hline
		5 & Life raft            &Map of the Moon's surface                      \\ \hline
	\end{tabular}
\end{table}

\subsection{Procedure}
\begin{figure}[t]
	\centering
	\includegraphics[width=\linewidth]{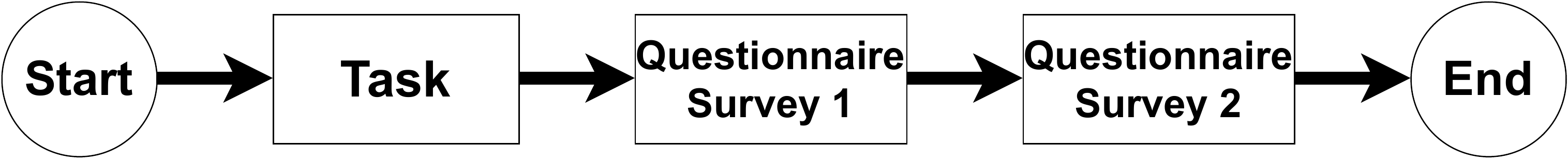}
	\caption{Flowchart of experiment}
	\label{fig:ex1}
\end{figure}

The procedure is shown in Fig. \ref{fig:ex1}. The participants viewed the instructions and watched one video from the four conditions. In the instructions, we stated that, ``In the experiment, you will watch a video of a human talking to a robot controlled by a human via remote control" and ``Watch as if you were talking to the robot." Afterward, we also told them that the task of the video was to have a discussion on moon survival. When participants finished watching the video, they were asked to rate their agreement on a seven-point Likert scale, 1 = strongly disagree, 7 = strongly agree, in two questionnaire surveys (in total, 30 statements). When they finished the experiment, there was an additional comment or question space. We paid 100 yen (about \$1.00US), and the average time to complete the study was about 15 to 30 minutes.
\subsection{Questionnaire Survey}
We used two different questionnaires, the Godspeed series and one for social presence. Godspeed is a standardized measurement tool for HRI \cite{Bartneck2009}. There are five key concepts for the measurement used: anthropomorphism, animacy, likeability, perceived intelligence, and perceived safety. For the second questionnaire, we used Networked Minds Measure of Social Presence, which is measure of presence \cite{Biocca2001}. We modified this questionnaire due to some of the statements not fitting into our experiment. A few of the questions from the first and second questionnaires are listed in Tables 1 and 2.
\begin{table}[ht]
	\centering
	\caption {GODSPEED's Questionnaire}
	\label{fig:gp}
	\begin{tabular}{|l|ll|}
		\hline
		& \textbf{ANTHROPOMORPHISM}           & \textbf{}        \\ \hline
		1 & \multicolumn{1}{l|}{Fake}           & Natural          \\ \hline
		2 & \multicolumn{1}{l|}{Machinelike}    & Humanlike        \\ \hline
		3 & \multicolumn{1}{l|}{Unconscious}    & Conscious        \\ \hline
		4 & \multicolumn{1}{l|}{Artificial}     & Lifelike         \\ \hline
		5 & \multicolumn{1}{l|}{Moving rigidly} & Moving elegantly \\ \hline
	\end{tabular}
\end{table}
\begin{table}[ht]
		\caption {Networked Minds Measure of Social Presence's Questionnaire}
		\label{fig:sp}
	\begin{tabular}{|l|l|ll}
		\cline{1-2}
		& \textbf{Social Presence}                                    & \textbf{} &  \\ \cline{1-2}
		1  & I often felt as if I was all alone                          &           &  \\ \cline{1-2}
		2  & I think the other individual often felt alone.              &           &  \\ \cline{1-2}
		3  & I was often aware of others in the environment.             &           &  \\ \cline{1-2}
		4  & Others were often aware of me in the room.                  &           &  \\ \cline{1-2}
		5  & The other individual paid close attention to me             &           &  \\ \cline{1-2}
		6  & I paid close attention to the other individual.             &           &  \\ \cline{1-2}
		7  & The other individual tended to ignore me.                   &           &  \\ \cline{1-2}
		8 & My behavior was in direct response to the other's behavior. &           &  \\ \cline{1-2}
		9 & etc.. &           &  \\ \cline{1-2}
	\end{tabular}
\end{table}
%
\section{result} 
To test our hypotheses, we used a two way analysis of variance (ANOVA). For the G*Power calculation  \cite{Erdfelder2009}, the sampling size was 128. For each condition, we used 32 participants for analysis.

Before participants watched a video, they were asked to rate their agreement on a seven-point Likert scale, 1 = not very familiar, 7 = very familiar, for two statements (``Are you familiar with video conferences?'' and ``Are you familiar with robots?''). Most participants were not familiar with robots (mean = 2.46, SD = 1.57 ), but they were familiar with video conferences (mean = 3.39, SD = 1.84). We used anthropomorphism from the GODSPEED questionnaire series to measure agency \cite{Bartneck2009}. To measure presence, we used Networked Minds Measure of Social Presence's questionnaire \cite{Biocca2001}. 
\subsection{Meausrement}
\begin{figure}[ht]
	\centering
	\includegraphics[width=9.0cm]{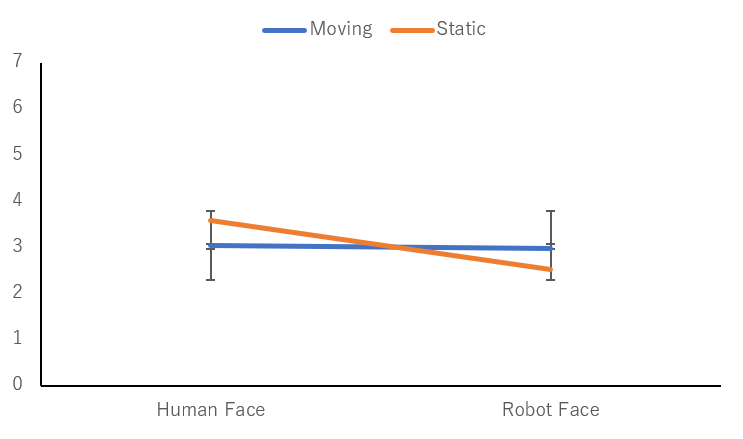}
	\caption{Averages for score for motion perceived for each condition in experiment. Anthropomorphism as dependent value.}
	\label{fig:ant}
\end{figure}

\begin{figure}[ht]
	\centering
	\includegraphics[width=9.0cm]{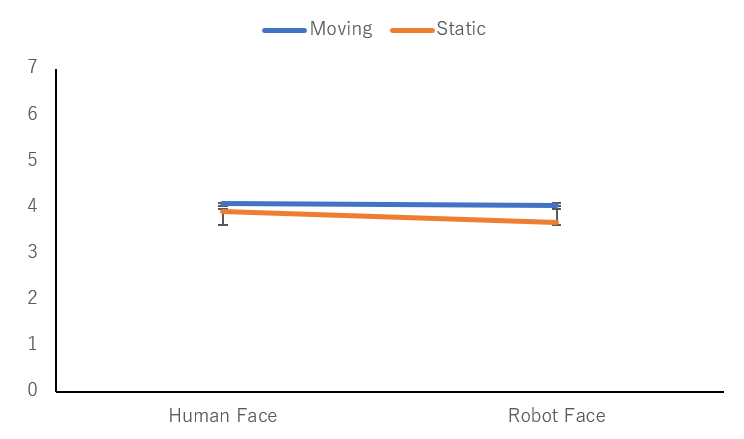}
	\caption{Averages for score for motion perceived for each condition in
		experiment. Presence as dependent value.}
	\label{fig:presence}
\end{figure}

The results of the ANOVA showed in Table \ref{fig:anova_ant} and Table \ref{fig:anovasp}, also result of simple mian effect for agency showed in Table \ref{fig:simple_effect_ant}. As well as the means and standard deviations(S.D.) for all of the dependent variables can be seen in Figs. \ref{fig:ant} and \ref{fig:presence} as well as in Table \ref{fig:antsd} and Table \ref{fig:presencesd}. Also, conditions explain again in Table \ref{fig:cond}. For agency, we found that the interaction was significant(p 	\textless 0.01). In the Static group, the simple main effect of face was significant(p \textless 0.0001). It was higher in the group with face. For presence, there was a significant interaction between the two factors. We found that the main effect was significant only for the motion factor(p \textless 0.01). Furthermore, motion with moving was the highest.
\begin{table}[tb]
	\centering
	\caption{Conditions in the experiments}
	\label{fig:cond}
	\begin{tabular}{lll}
		\hline
		\textbf{condition} & \textbf{Face} & \textbf{Motion} \\ \hline
		condition1         & Human Face    & Moving          \\
		condition2         & Human Face    & Static          \\
		condition3         & Robot Face    & Moving          \\
		condition4        & Robot Face    & Static          \\ \hline
	\end{tabular}
\end{table}
%
\begin{table}[ht]
	\centering
	\caption{Agency of S.D. and Mean}
	\label{fig:antsd}
	\begin{tabular}{lll}
		\hline
		\textbf{Condition} & \textbf{MEAN} & \textbf{S.D.} \\ \hline
		condition1         & 3.05          & 0.92          \\
		condition2         & 3.57          & 1.34          \\
		condition3         & 2.98          & 0.91          \\
		condition4         & 2.51          & 1.08          \\ \hline
	\end{tabular}
\end{table}
%
\begin{table}[ht]
	\centering
	\caption{Presence of S.D. and Mean}
	\label{fig:presencesd}
	\begin{tabular}{lll}
		\hline
		\textbf{Condition} & \textbf{MEAN} & \textbf{S.D.} \\ \hline
		condition1         & 4.08          & 0.54          \\
		condition2         & 3.91          & 0.42          \\
		condition3         & 4.03          & 0.52          \\
		condition4         & 3.66          & 0.67          \\ \hline
	\end{tabular}
\end{table}
\begin{table}[ht]
	\centering
	\caption{RESULT OF TWO-WAY ANOVA FOR Presence}
	\label{fig:anovasp}
\begin{tabular}{llll}
	\hline
	\textbf{Source} & \textbf{F(1,124)} & \textbf{p} &    \\ \hline
	Face            & 2.457             & 0.119      & n.s. \\
	Motion          & 8.144             & 0.005      & ** \\
	Interaction     & 1.00              & 0.316      & n.s. \\ \hline
\end{tabular}
\begin{tablenotes}\footnotesize
	\centering
	\item[*]  *p \textless .05,**p \textless 0.01 
\end{tablenotes}
\end{table}
\begin{table}[ht!]
	\centering
	\caption{RESULT OF TWO-WAY ANOVA FOR Agency}
	\label{fig:anova_ant}
	\begin{tabular}{llll}
		\hline
		\textbf{Source} & \textbf{F(1,124)} & \textbf{p} &    \\ \hline
		Face            & 8.776             & 0.003      & ** \\
		Motion          & 0.021             & 0.883      & n.s. \\
		Interaction     & 6.772             & 0.010      & *  \\ \hline
	\end{tabular}
	\begin{tablenotes}\footnotesize
		\centering
		\item[*]  *p \textless .05,**p \textless 0.01 
	\end{tablenotes}
\end{table}
\begin{table}[ht!]
	\centering
	\caption{RESULT OF SIMPLE MAIN EFFECT TEST FOR Agency}
	\label{fig:simple_effect_ant}
\begin{tabular}{llll}
	\hline
	\textbf{Simple Main Effect} & \textbf{F(1,124)} & \textbf{p} &     \\ \hline
	Face (Moving)               & 0.0648            & 0.7994     & n.s.  \\
	Face (Static)               & 15.4846           & 0.0001     & *** \\ \hline
	Motion (Human Face)         & 3.7806            & 0.0541     & n.s.   \\
	Motion (Robot Face)         & 3.0139            & 0.0850     & n.s.   \\ \hline
\end{tabular}
	\begin{tablenotes}\footnotesize
		\centering
		\item[*]  *p \textless .05,**p \textless 0.01,***p \textless 0.001
	\end{tablenotes}
\end{table}
\section{Discussion}
\subsection{Hypotheses Summary}
The first section in the Experiment section set four hypotheses for what we expected to be the outcome of the experiment. Here are the results:
\begin{itemize}
	\item \textbf{H1} Face affects agency: In the Static group, a simple main effect was found for the face factor (p \textless 0.0001), which was partially supported.
	\item \textbf{H2} Motion affects agency: This hypothesis did not hold since a simple main effect was not found for the motion factor.
	\item \textbf{H3} Face affects presence: This hypothesis did not hold since there was no main effect of the face factor on presence.
	\item \textbf{H4} Motion affects presence: This hypothesis held since a main effect on presence was found for the motion factor (p \textless 0.01).
\end{itemize}
When the telepresence robot had a Human Face and no arm motion which is Static, the participants felt the robot to be human-like. Since a real human's face could be seen, they felt this robot to be most human-like. Even when the human's face appeared with motion which is moving, the participants did not feel the robot to be not human-like.
Regarding the presence factor, participants felt presence when there was robot do motions since main effect on presence factor (p \textless 0.01). So, it doesn't matter with between the human face and robot-like face by reason of no main effect on face factor.

From additional comments and questions from the survey, several participants mentioned that it was difficult to watch the video and answer questions because the motors were too loud, but there was no problem with the motions. 
\subsection{Generality and Limitation}
There are certain generalities and the study presented here has limitations that may affected the generalities of our result. First, our use of a Rapiro as robot platform might affect generality since we only tested this with a particular two robots. Naturally, each type of robot embodiments had own limitations due to number of actuators. However, the finding of robot embodiment could not necessarily be generalized to other types of robotic embodiments \cite{Mollahosseini2018}. Second, the lack of support for hypotheses in motion what we developed robot's motion which used in experiments and motion lists showed in Table \ref{fig:motionlist}. In addition, the delay between remote operator's speaking and motion playback speed also type of motion might affect perceptions of the presence and agency. 

Third, Robot face may affected generality. We already mention above, the finding of robot embodiment could not necessarily be generalized to other types of robotic embodiments \cite{Mollahosseini2018} and we believed also same as facial cue. 
In the other hand, Japanese people often imagine humanoid robots as friendly \cite{Yamaoka2007}, since we only used Japanese participants, unsure about people in other culture can be applied in our results. 

Regarding online experiments in general, Crump et al. \cite{pone} showed that the data collected online using a web-browser seemed mostly in line with laboratory results, so long as the experimental methods were solid.

There were problems with the human-likeness and machine-likeness from the video task. When the face was robot-like, for even those participants not used to robots, the robot appeared human-like, and the way the robot talked sounded human-like.
When the face was that of a human, the participants felt the robot to be uncanny because it had a human face but a robot body. This may relate to the uncanny valley. Furthermore, most participants felt as if they were having a video chat for the condition using a Human Face without arm motion which is Static. In future work, we will conduct in-person experiment same conditions and method in this study. Also, we will compare between video chat and this condition of using a human face and no motion. 

Last, we found that no matter which robot we used, even simple motions could invoke a feeling of presence. Regarding human-face conditions, there was no significant effect of gender in Sirkin et al. \cite{Sirkin2012}. The design of future telepresence robots will change depending on whether the emphasis is on human likeness or presence.
\section{CONCLUSIONS}
In prior research, Non-verbal behaviors and facial cues are interesting and essential cues in robot and computer-mediated communication.  The present study showed that individuals study that non-verbal behavior and facial cue in the telepresence robot's study. The result from this study advances our understanding role of the nonverbal and facial cues in robot-mediated communication.  

In this study, we conducted $2\times2$ between-participant experiment(robot face: video-conference, robot-like face; arm motion: move, static). The result showed that people felt a presence at motion factor even though doesn't matter with the human face and robot-like face. Furthermore, anthropomorphism as the agency, people felt more in human face with static which is more human-like than other conditions. 

A comparison that only video-conference and human face embodiment with moving or static will be future work for us. While the quality of communication through embodied web-based experiment may never same as in-person experiment, however our study facial and motion cue could support and helped social interaction using a telepresence robot.



%
%
%
%


\bibliographystyle{IEEEtran.bst}
\bibliography{main.bib}

\begin{thebibliography}{10}
\providecommand{\url}[1]{#1}
\csname url@rmstyle\endcsname
\providecommand{\newblock}{\relax}
\providecommand{\bibinfo}[2]{#2}
\providecommand\BIBentrySTDinterwordspacing{\spaceskip=0pt\relax}
\providecommand\BIBentryALTinterwordstretchfactor{4}
\providecommand\BIBentryALTinterwordspacing{\spaceskip=\fontdimen2\font plus
\BIBentryALTinterwordstretchfactor\fontdimen3\font minus
  \fontdimen4\font\relax}
\providecommand\BIBforeignlanguage[2]{{%
\expandafter\ifx\csname l@#1\endcsname\relax
\typeout{** WARNING: IEEEtran.bst: No hyphenation pattern has been}%
\typeout{** loaded for the language `#1'. Using the pattern for}%
\typeout{** the default language instead.}%
\else
\language=\csname l@#1\endcsname
\fi
#2}}

\bibitem{Kristoffersson2013}
A.~Kristoffersson, S.~Coradeschi, and A.~Loutfi, ``{A review of mobile robotic
  telepresence},'' \emph{Advances in Human-Computer Interaction}, vol. 2013,
  2013.

\bibitem{Rae2013}
I.~Rae, L.~Takayama, and B.~Mutlu, ``{The influence of height in robot-mediated
  communication},'' \emph{ACM/IEEE International Conference on Human-Robot
  Interaction}, pp. 1--8, 2013.

\bibitem{Rae2014}
I.~Rae, B.~Mutlu, and L.~Takayama, ``{Bodies in motion},'' in \emph{Proceedings
  of the 32nd annual ACM conference on Human factors in computing systems - CHI
  '14}.\hskip 1em plus 0.5em minus 0.4em\relax New York, New York, USA: ACM
  Press, 2014, pp. 2153--2162.

\bibitem{Lee2011}
M.~K. Lee and L.~Takayama, ``{"Now, i have a body"},'' p.~33, 2011.

\bibitem{Tsui2011}
K.~M. Tsui, M.~Desai, H.~A. Yanco, and C.~Uhlik, ``{Exploring use cases for
  telepresence robots},'' \emph{HRI 2011 - Proceedings of the 6th ACM/IEEE
  International Conference on Human-Robot Interaction}, no. August, pp. 11--18,
  2011.

\bibitem{Neustaedter2016}
C.~Neustaedter, G.~Venolia, J.~Procyk, and D.~Hawkins, ``{To beam or not to
  beam: A study of remote telepresence attendance at an academic conference},''
  \emph{Proceedings of the ACM Conference on Computer Supported Cooperative
  Work, CSCW}, vol.~27, pp. 418--431, 2016.

\bibitem{Fels2001}
D.~I. Fels, J.~K. Waalen, S.~Zhai, and P.~T. Weiss, ``{Telepresence under
  exceptional circumstances: enriching the connection to school for sick
  children},'' \emph{Proceedings of Interact}, no. September 2015, pp.
  617--624, 2001.

\bibitem{movement17}
M.~Choi, R.~Kornfield, L.~Takayama, and B.~Mutlu, ``Movement matters: Effects
  of motion and mimicry on perception of similarity and closeness in
  robot-mediated communication,'' pp. 325--335, 2017.

\bibitem{Takeuchi2020}
K.~Takeuchi, Y.~Yamazaki, and K.~Yoshifuji, ``Avatar work: Telework for
  disabled people unable to go outside by using avatar robots,'' p. 53–60,
  2020.

\bibitem{Adalgeirsson2010}
S.~O. Adalgeirsson and C.~Breazeal, ``{MeBot: A robotic platform for socially
  embodied presence},'' \emph{5th ACM/IEEE International Conference on
  Human-Robot Interaction, HRI 2010}, pp. 15--22, 2010.

\bibitem{Orylab}
``Orihime,'' \url{https://orylab.com/en/\#product}.

\bibitem{McGinn2020a}
C.~McGinn, E.~Bourke, A.~Murtagh, C.~Donovan, P.~Lynch, M.~F. Cullinan, and
  K.~Kelly, ``{Meet Stevie: a Socially Assistive Robot Developed Through
  Application of a ‘Design-Thinking' Approach},'' \emph{Journal of
  Intelligent and Robotic Systems: Theory and Applications}, vol.~98, no.~1,
  pp. 39--58, 2020.

\bibitem{McGinn2020}
C.~McGinn, ``{Why Do Robots Need a Head? The Role of Social Interfaces on
  Service Robots},'' \emph{International Journal of Social Robotics}, vol.~12,
  no.~1, pp. 281--295, 2020.

\bibitem{Mollahosseini2018}
A.~Mollahosseini, H.~Abdollahi, T.~D. Sweeny, R.~Cole, and M.~H. Mahoor,
  ``{Role of embodiment and presence in human perception of robots' facial
  cues},'' \emph{International Journal of Human Computer Studies}, vol. 116,
  no. April, pp. 25--39, 2018.

\bibitem{Vikkels2020}
\BIBentryALTinterwordspacing
S.~Vikkels{\o}, T.~H. Hoang, F.~Carrara, K.~D. Hansen, and B.~Dinesen, ``{The
  telepresence avatar robot OriHime as a communication tool for adults with
  acquired brain injury: an ethnographic case study},'' \emph{Intelligent
  Service Robotics}, vol.~13, no.~4, pp. 521--537, 2020. [Online]. Available:
  \url{https://doi.org/10.1007/s11370-020-00335-6}
\BIBentrySTDinterwordspacing

\bibitem{Kuzuoka2010}
H.~Kuzuoka, Y.~Suzuki, J.~Yamashita, and K.~Yamazaki, ``{Reconfiguring spatial
  formation arrangement by robot body orientation},'' \emph{5th ACM/IEEE
  International Conference on Human-Robot Interaction, HRI 2010}, pp. 285--292,
  2010.

\bibitem{Yamaoka2007}
F.~Yamaoka, T.~Kanda, H.~Ishiguro, and N.~Hagita, ``{Interacting with a human
  or a humanoid robot?}'' \emph{IEEE International Conference on Intelligent
  Robots and Systems}, pp. 2685--2691, 2007.

\bibitem{Yamada2013}
S.~Yamada, T.~Komatsu, K.~Terada, K.~Funakoshi, K.~Kobayash, and M.~Nakano,
  ``{Expressing a Robot's Confidence with Motion-based Artificial Subtle
  Expressions},'' \emph{Conference on Human Factors in Computing Systems -
  Proceedings}, vol. 2013-April, pp. 1023--1028, 2013.

\bibitem{Sirkin2012}
D.~Sirkin and W.~Ju, ``{Consistency in physical and on-screen action improves
  perceptions of telepresence robots},'' \emph{HRI'12 - Proceedings of the 7th
  Annual ACM/IEEE International Conference on Human-Robot Interaction}, pp.
  57--64, 2012.

\bibitem{rapiro}
``Rapiro,'' \url{http://www.rapiro.com/ja/}.

\bibitem{Abiri2015}
R.~Abiri, J.~McBride, X.~Zhao, and Y.~Jiang, ``{A real-time brainwave based
  neuro-feedback system for cognitive enhancement},'' \emph{ASME 2015 Dynamic
  Systems and Control Conference, DSCC 2015}, vol.~1, no. September 2016, 2015.

\bibitem{DAuria2018}
D.~D'Auria, B.~Siciliano, F.~Persia, F.~Bettini, and S.~Helmer, ``{SARRI: A
  SmArt Rapiro robot integrating a framework for automatic high-level
  surveillance event detection},'' \emph{Proceedings - 2nd IEEE International
  Conference on Robotic Computing, IRC 2018}, vol. 2018-Janua, pp. 238--241,
  2018.

\bibitem{Erdfelder2009}
E.~Erdfelder, F.~FAul, A.~Buchner, and A.~G. Lang, ``{Statistical power
  analyses using G*Power 3.1: Tests for correlation and regression analyses},''
  \emph{Behavior Research Methods}, vol.~41, no.~4, pp. 1149--1160, 2009.

\bibitem{Powers2007}
A.~Powers, S.~Kiesler, S.~Fussell, and C.~Torrey, ``{Comparing a computer agent
  with a humanoid robot},'' \emph{HRI 2007 - Proceedings of the 2007 ACM/IEEE
  Conference on Human-Robot Interaction - Robot as Team Member}, pp. 145--152,
  2007.

\bibitem{Woods2006}
S.~N. Woods, M.~L. Walters, K.~L. Koay, and K.~Dautenhahn, ``{Methodological
  issues in HRI: A comparison of live and video-based methods in robot to human
  approach direction trials},'' \emph{Proceedings - IEEE International Workshop
  on Robot and Human Interactive Communication}, pp. 51--58, 2006.

\bibitem{lafferty74}
J.~C. Lafferty, Eady, and J.~Elmers, ``{The desert survival problem},'' 1974.

\bibitem{moon}
``Nasa exercise: Ranking survival objects for the moon,''
  \url{https://www.psychologicalscience.org/observer/nasa-exercise}.

\bibitem{Biocca2001}
F.~Biocca, C.~Harms, and J.~Gregg, ``{The Networked Minds Measure of Social
  Presence : Pilot Test of the Factor Structure and Concurrent Validity
  Co-Presence},'' 2001.

\bibitem{Yonezu2017}
S.~Yonezu and H.~Osawa, ``{Telepresence robot with behavior synchrony: Merging
  the emotions and behaviors of users},'' \emph{RO-MAN 2017 - 26th IEEE
  International Symposium on Robot and Human Interactive Communication}, vol.
  2017-Janua, pp. 213--218, 2017.

\bibitem{Bartneck2009}
C.~Bartneck, D.~Kuli, and E.~Croft, ``{Measurement Instruments for the
  Anthropomorphism , Animacy , Likeability , Perceived Intelligence , and
  Perceived Safety of Robots},'' pp. 71--81, 2009.

\bibitem{pone}
M.~J.~C. Crump, J.~V. McDonnell, and T.~M. Gureckis, ``Evaluating amazon's
  mechanical turk as a tool for experimental behavioral research,'' \emph{PLOS
  ONE}, vol.~8, no.~3, pp. 1--18, 03 2013.

\end{thebibliography}
\end{document}